\documentclass[a4paper]{article}

\usepackage{INTERSPEECH2019}

\usepackage{graphicx}
\usepackage{amsmath}
\usepackage{graphicx}
\usepackage{hyperref}
\usepackage{multirow}
\usepackage{subfigure}

\title{Unsupervised low-rank representations for speech emotion recognition}
\name{Georgios Paraskevopoulos$^{1, 2}$ \quad Efthymios Tzinis$^3$ \quad Nikolaos Ellinas$^1$ \\ Theodoros Giannakopoulos$^2$ \quad Alexandros Potamianos$^{1, 2}$}
\address{
  $^1$School of Electrical \& Computer Engineering, National Technical University of Athens, Greece\\
  $^2$Behavioral Signal Technologies, Los Angeles, CA, USA\\
  $^{3}$Department of Computer Science, University of Illinois at Urbana-Champaign, IL, US}

\email{geopar@central.ntua.gr, etzinis@gmail.com, nellinas@central.ntua.gr, thodoris@behavioralsignals.com, potam@central.ntua.gr}

\begin{document}

\maketitle
\begin{abstract}
We examine the use of linear and non-linear dimensionality reduction algorithms for extracting
low-rank feature representations for speech emotion recognition. Two feature
sets are used, one based on low-level descriptors and their aggregations (IS10) and one modeling recurrence dynamics of speech (RQA), as well as their fusion. 
We report speech  emotion recognition (SER) results for learned representations on two databases using
different classification methods. Classification with low-dimensional representations
yields performance improvement in a variety of settings. This indicates that dimensionality reduction is an effective way to combat the curse of dimensionality for SER. Visualization of features in two dimensions provides insight into discriminatory abilities of reduced feature sets.

\end{abstract}

\section{Introduction}
\label{sec:intro}
Human-machine interaction is constantly evolving towards the use of more natural interfaces, like speech. Still the key difference between human-human and human-machine communication is the ability of humans to recognize the emotion of their conversation peers and modify their communication strategy based on that. 
Although significant progress has been done in the field of speech emotion recognition (SER), machines have not achieved human-like performance. 
One of the reasons is the scarcity of available annotated data. SER databases are mostly composed of relatively small number of utterances from few speakers, which limits the generalization abilities of the models. Furthermore modern SER systems rely on feature sets of high dimensions. The small amount of training samples do not cover all combinations of values in the high-dimensional feature spaces and, thus, SER algorithms suffer from the curse of dimensionality (CoD) \cite{bellman2015adaptive}. 
In this work we postulate that reducing  dimensionality of the feature space is an effective way to combat CoD and demonstrate that low-dimensional representations yield simpler models with comparable performance.
Dimensionality reduction (DR) algorithms aim at learning low-dimensional latent representations of real world data. Such representations can be used for exploratory data analysis, to visualize and gain intuition on the statistical properties of data  or, as in our case, extract latent features for input to classification or regression models. 

Evidence that DR on speech features can create robust representations for SER can be found in the literature.
In \cite{chiou2013feature}, Principal Component Analysis (PCA) \cite{Pearson} is used to extract low-dimensional representations for the feature set introduced in \cite{schuller2009acoustic} containing $6552$ features. The system is evaluated on Berlin emotional database (Emo-DB) \cite{burkhardt2005database}.
In \cite{yuan2015dimension} Linear Discriminant Analysis (LDA) \cite{fisher1936use} and PCA are used for SER, along with a weighted variation of LDA on a feature set of $225$ dimensions. Experiments showed no significant performance difference between PCA and LDA.
These methods are also compared in \cite{you2006emotion} and \cite{you2006hierarchical}, along with Sequential Forward Selection (SFS) \cite{fu1968sequential}, on a feature set consisting of $48$ prosodic features and $16$ formants.
PCA representations extracted in \cite{you2006emotion} are found to be inferior than LDA, while \cite{you2006hierarchical} observed no significant difference.
SFS and PCA are also explored in \cite{ververidis2004automatic} for the Danish Emotion Speech database \cite{engberg1997design}. 
\cite{lee2002classifying} experimentally found that applying PCA on utterance-level statistics of pitch and energy features gives equivalent SER performance with the original features on a call center dialog corpus. 
Authors in \cite{chuang2004emotion} report that classification accuracy keeps improving when increasing the number of principal components only up to a centain rank for a feature set of $33$ dimensions.
A supervised variation of PCA along with Greedy Feature Selection (GFS) \cite{farahat2011efficient} and ElasticNet \cite{friedman2010regularization} are explored in \cite{fewzee2012dimensionality} on two sets of energy-based and MFCC-based feature sets of $400$ and $82$ dimensions, with inconclusive results as to which approach is superior.
The application of Linear and non-linear DR methods on SER is examined on a prosodic feature set of $48$ dimensions in \cite{zhang2013dimensionality}. Compared methods include unsupervised methods like PCA, Isometric Mapping (ISOMAP) \cite{tenenbaum_global_2000} and Locally Linear Embedding (LLE) \cite{roweis2000nonlinear}, and supervised methods LDA, Supervised LLE (SLLE) \cite{de2003supervised}, Neighborhood Component Analysis (NCA) \cite{goldberger2005neighbourhood}, Maximally Component Metric Learning (MCML) \cite{globerson2006metric}, local Fisher Discriminant Analysis (LFDA) \cite{sugiyama2007dimensionality} and Modified SLLE (MSLLE). Results show better performance of PCA for unsupervised DR while MSLLE was superior for supervised DR.



\section{Dimensionality Reduction Algorithms}
\label{sec:manifold}

DR algorithms compress data in a low-dimensional space while preserving meaningful statistical and geometrical properties. Such properties are covariance of original data, pairwise distances between samples or local neighborhoods.
They can be separated into two general categories, linear and non-linear. Linear DR aims to find a linear projection $Y=TX \in R^{n \times k}$  of the real data $X \in R^{n \times m}$, where $k < m$. Examples of linear DR algorithms are PCA and classical multidimensional scaling (cMDS) \cite{torgerson1952multidimensional}. PCA projects data into a low-dimensional space, which is formed by an orthogonal basis of linearly uncorrelated vectors called the principal components. Principal components are selected as the axes along which the samples have maximum variance. cMDS takes a geometric approach, finding a set of low-dimensional points that best preserve pairwise euclidean distances between original data points. 

Non-linear dimensionality reduction (NLDR) algorithms aim to infer the intrinsic geometry of the original data, based on the manifold hypothesis, which states that real world data tend to lie on a low-dimensional manifold, embedded in the high-dimensional space. These algorithms are not limited in linear transformations, like the rotations and stretches that can be induced by a matrix multiplication.
An extension of cMDS is metric MDS \cite{kruskal1964multidimensional} where dissimilarity measures are assumed metric, but not necessarily euclidean. When these measures are closely related to the euclidean distance, e.g. cosine distance, metric MDS is still characterized as a linear DR approach. Stress Majorization \cite{kruskal1964multidimensional} and Pattern Search MDS \cite{paraskevopoulosPSMDS} are two algorithms for metric MDS.
The non-metric extension of MDS \cite{kruskal1964nonmetric} tries to approximate the rank order of original distances by applying a monotonically increasing function, usually approximated by isotonic regression. ISOMAP finds an isometric mapping of the original data by extending metric MDS to approximate geodesic pairwise distances between original samples space as euclidean pairwise distances in the transformed samples. Geodesic distances are approximated by the shortest path distances between data points. While MDS and ISOMAP consider the global data geometry, Local Linear Embedding (LLE) reconstructs local regions by finding sets of weights which are used to represent samples as a weighted combination of their closest neighbors. Representations are computed by solving a sparse eigenvalue problem. Modified LLE \cite{zhang2007mlle} is an extension of LLE that uses multiple neighborhood weights and produces more robust results.
Another non-linear approach is Laplacian Eigenmaps or Spectral Embedding \cite{belkin2002laplacian}, which preserves local manifold geometry by minimizing the Laplacian of the graph formed by neighboring data points. The Laplacian of this graph approximates the Laplacian-Beltrami operator over the manifold, which indicates the divergence of the mapping of a high-dimensional point to the low-dimensionsional manifold.

Autoencoders \cite{ballard1987modular} are a class of deep neural networks that can be used for linear and non-linear dimensionality reduction and are composed of an encoder and a decoder. Encoder projects input $x$ to a low-dimensional space via a hidden layer $h$, while the attempts to reconstruct $x$ from $h$. If no non-linear activations are used, encoder learns a linear projection $W x + b$, whereas if we use a non-linear activation function (e.g. sigmoid or rectified linear unit) in the output of the encoder's layers, a non-linear embedding is learned.

\begin{figure*}[htb!]
\centering
\subfigure[RQA feature set on Emo-DB\label{fig:rqa:emodb}]{
    \includegraphics[width=.32\textwidth]{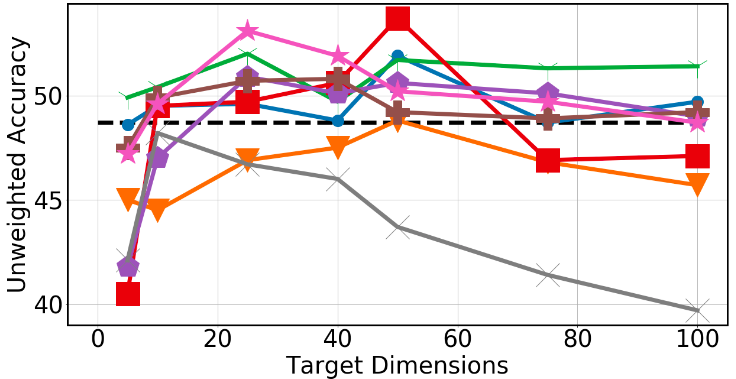}
}
\subfigure[IS10 feature set on Emo-DB\label{fig:is10:emodb}]{
  \includegraphics[width=.32\textwidth]{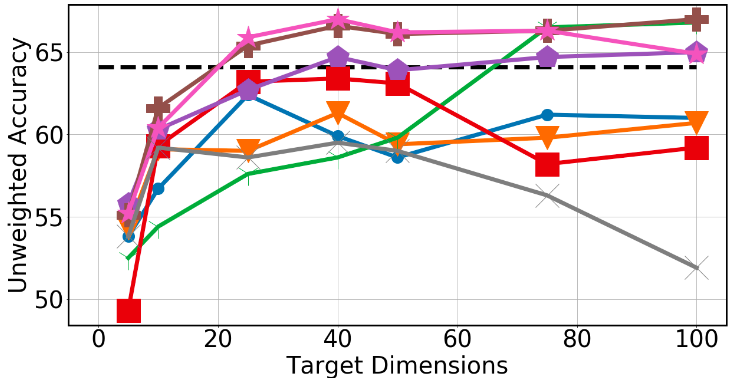}
}
\subfigure[Fused feature set on Emo-DB\label{fig:fused:emodb}]{
  \includegraphics[width=.32\textwidth]{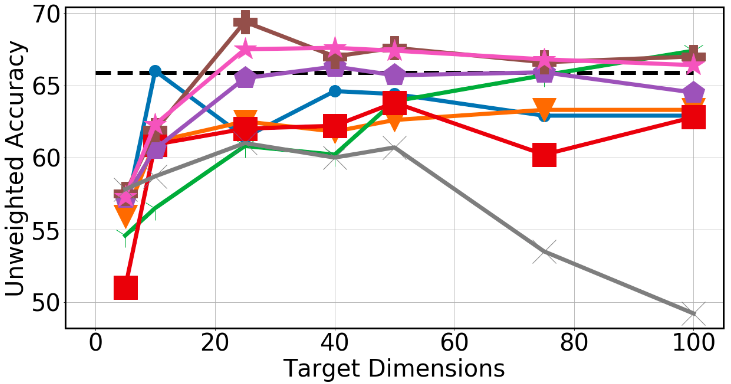}
}
\subfigure[RQA feature set on IEMOCAP\label{fig:rqa:iemocap}]{
    \includegraphics[width=.32\textwidth]{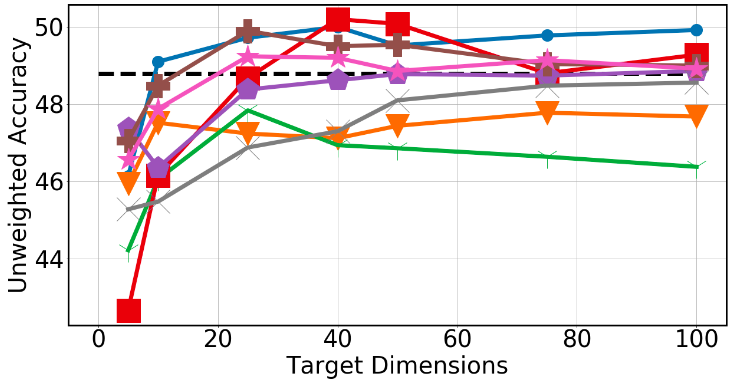}
}
\subfigure[IS10 feature set on IEMOCAP\label{fig:is10:iemocap}]{
  \includegraphics[width=.32\textwidth]{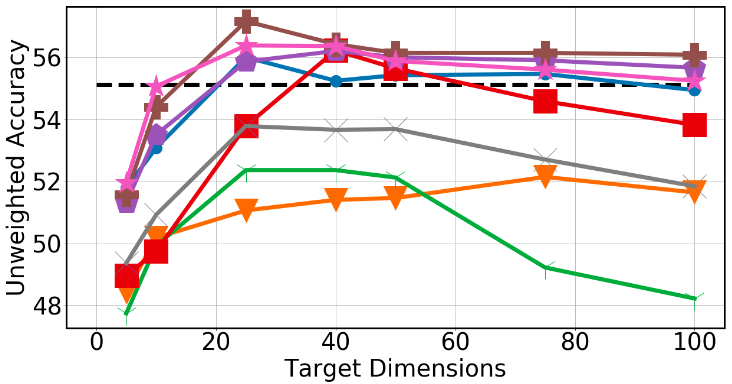}
}
\subfigure[Fused feature set on IEMOCAP\label{fig:fused:iemocap}]{
  \includegraphics[width=.32\textwidth]{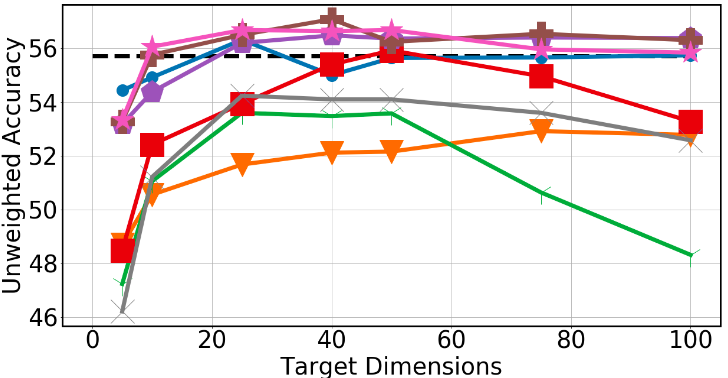}
}
\subfigure{
  \includegraphics[width=.5\textwidth]{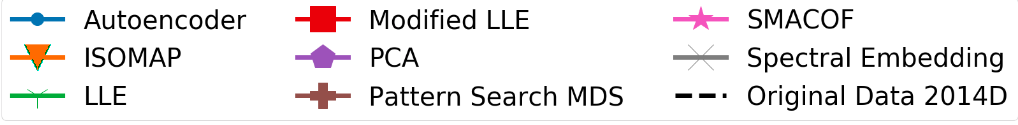}
}

\caption{Results of DR for different feature sets on IEMOCAP and Emo-DB}





\end{figure*}

\section{Features for speech emotion recognition}
\label{sec:features}

We consider the following feature sets:\vspace*{1mm}

\noindent \textbf{IS10 set}: The IS10 feature set \cite{schuller2010interspeech} consists of $1582$ features. IS10 is obtained by transforming the signal in the Fourier space. Features correspond to $21$ statistical functionals (e.g. percentiles, linear regression coefficients) applied to $38$ low level descriptors (MFCCs, PCM loudness etc.) and their deltas. Extraction is performed using the openSMILE toolkit.\vspace*{1mm}

\noindent \textbf{RQA set}: The Recurrence Quantification Analysis (RQA) feature set \cite{tzinis2018integrating} consists of $432$ features. This feature set is obtained by analyzing speech dynamics through phase space representation. The phase space is reconstructed through the use of time-delayed versions of the original signal and then the recurrence plots are calculated as thresholded pairwise distances of points in the phase space. Features are extracted as aggregated RQA measures from the recurrence plots.
Source code for feature extraction is publicly available.\footnote{\href{https://github.com/etzinis/nldrp}{https://github.com/etzinis/nldrp}}\vspace*{1mm}

\noindent \textbf{Fused set}: We concatenate features from IS10 and RQA into a represenation of $2014$ dimensions, modeling both frequency content of speech signals and recurrence dynamics.










\section{Experiments and Results}
\label{sec:expres}

\subsection{Experimental Setup}
\label{ssec:expset}
We use the following databases for evaluation:\vspace*{1mm}

\noindent\textbf{Emo-DB:} Berlin Database of Emotional Speech (Emo-DB) \cite{burkhardt2005database} contains $535$ emotional German sentences, voiced by $10$ actors ($5$ male and $5$ female). Specifically, $7$ emotions are included i.e., $127$ anger, $45$ disgust, $70$ fear, $71$ joy, $60$ sadness, $81$ boredom and $70$ neutral.\vspace{1mm}

\noindent\textbf{IEMOCAP:} IEMOCAP database \cite{busso2008iemocap} contains $12$ hours of video data with scripted and improvised dialog recorded by $10$ actors. Utterances are organized in $5$ sessions of dyadic interactions between pairs of actors. For our experiments we consider $5531$ utterances of 4 emotions (1103 angry, 1636 happy, 1708 neutral and 1084 sad), where we merge excitement class into happiness \cite{aldeneh2017using}, \cite{xia2017multi}, \cite{fayek2017evaluating}, \cite{ghosh2016representation}. \vspace*{1mm}

We consider utterance-level, speaker independent (SI) SER for our experiments. In this setup a number of speakers are kept hidden from the training set and used for evaluation. Specifically in the case of Emo-DB we perform leave one speaker out (LOSpO) cross-validation, where test folds contain the instances of the unknown speaker. For IEMOCAP we use the leave one session out (LOSO) cross-validation scheme, where two speakers participating in a session are used as the evaluation folds. This results in a $10$-fold cross-validation scheme for Emo-DB and $5$-fold cross-validation for IEMOCAP. We apply $Z$-normalization to standardize the features in zero mean and unit variance, where each sample $x$ is transformed according to the formula $z = \frac{x - \mu}{\sigma}$.
 Note that for SI experiments only samples in the training set are used to calculate $\mu$ and $\sigma$ and test samples are normalized using these statistics.
 
 Representations resulting from all DR approaches are evaluated for $k$-nearest neighbors ($k$NN) classification. We perform grid search on the optimal number of neighbors $k$ in the $[1, 30]$ range and report results for the optimal value for each dimension and each method. Optimal values of $k$ range from $13$ to $20$ indicating that consistent neighborhoods are formed in the low-dimensional spaces. We also evaluate low-rank representations on SVM with linear and gaussian kernels, and Logistic Regression (LR), with optimal value of $C$ in the range $[0.01, 10]$.
 Autoencoder is trained with $3$ encoder layers, $3$ decoder layers and $1$ hidden layer, using ReLU activations.
 
\subsection{Results}
\label{sec:res}

As evaluation metrics we used both weighted accuracy and unweighted accuracy. For brevity we report unweighted accuracy results, noting that same trends form with respect to the weighted accuracy metric. Fig.~\ref{fig:rqa:emodb} shows the results of DR applied to the RQA features on Emo-DB for all DR methods, for different embedding dimensions $L$. We observe that Modified LLE achieves the best results when $L=50$, followed by SMACOF MDS in $L=25$. Observe that all methods except ISOMAP and Spectral Embedding manage to outperform the original features of $432$ dimensions. In Fig.~\ref{fig:is10:emodb}, which shows results for DR on IS10 features for Emo-DB, we can observe a different pattern. Here the MDS algorithms perform best for every embedding dimension, followed by PCA, all three of these methods outperforming the original feature set of $1582$ dimensions. This indicates that this feature set resembles more a hyperplane in the high-dimensional space than a non-linear manifold. Non-linear methods like LLE, ISOMAP and Spectral Embedding underperform. For the fused feature set in Fig.~\ref{fig:fused:emodb} we see again that distance-preserving transformations yield the best performance.
Same patterns emerge in IEMOCAP in Fig.~\ref{fig:rqa:iemocap},~\ref{fig:is10:iemocap},~\ref{fig:fused:iemocap}, with Modified LLE achieving better performance for the RQA features and Pattern Search MDS and PCA yielding best representations for IS10 features. Notably in IEMOCAP, performance of the Autoencoder is significantly better because there are more training samples. For the experiments with the fused feature set we again observe a consistent pattern in both Emo-DB and IEMOCAP, with MDS yielding again the best representations followed by PCA. Fusion is still beneficial after applying DR though we observe that the structure of IS10 features dominates under fusion.

In Table~\ref{tab:res} we show results for linear SVM, radial basis function (rbf) SVM, $k$NN and LR. We reduce dimensionality of IS10 features from $1582$ to $25$ dimensions and report unweighted accuracy (UA) on IEMOCAP. Low-rank representations produce very competitive results to the original sparse features, while for linear SVM and $k$NN they even improve classification accuracy. Overall global, linear DR methods like MDS and PCA produce the best representations.

\subsection{Visualization}
\label{sec:vis}

We include visualizations of feature maps reduced in $2$D. We focus on the best and the worst performing methods and comment on some interesting observations.

\begin{figure*}
\centering
\subfigure[Series\label{fig:cross:series}]{
    \includegraphics[width=.32\textwidth]{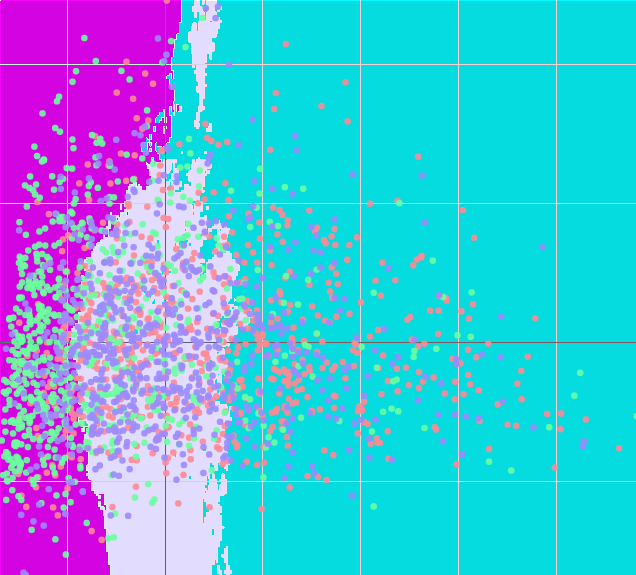}
}
\subfigure[Movies\label{fig:cross:movies}]{
    \includegraphics[width=.32\textwidth]{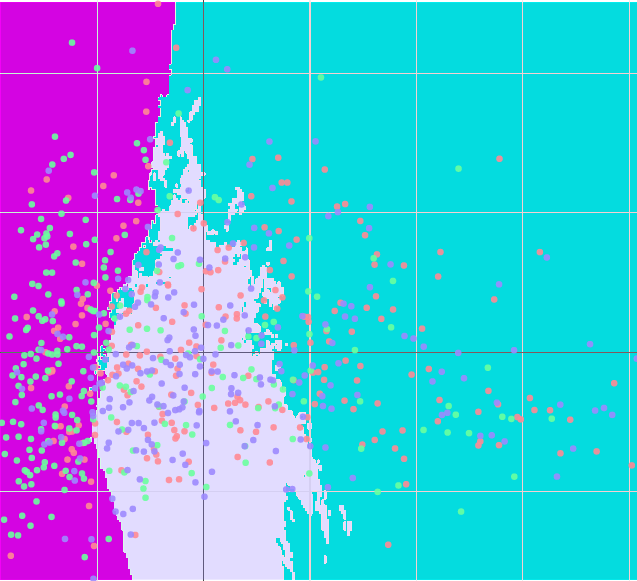}
}
\subfigure[Interviews\label{fig:cross:interviews}]{
    \includegraphics[width=.32\textwidth]{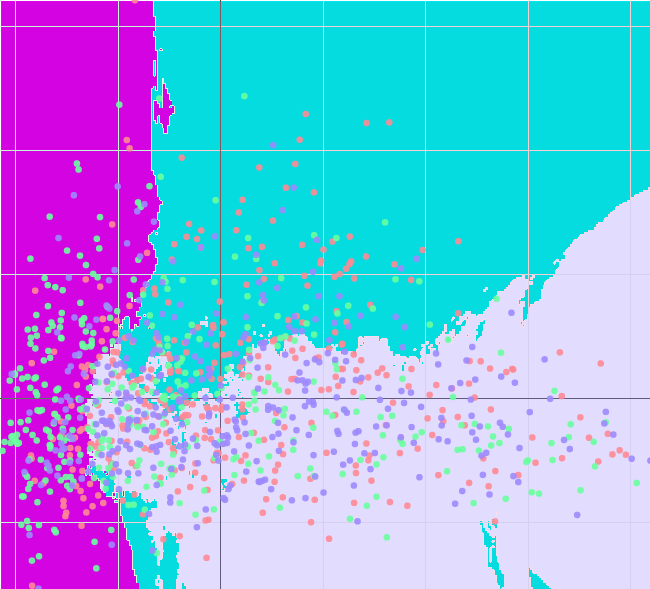}
}
\caption{Cross-domain decision regions with $2$D DR}\label{fig:cross}
\end{figure*}

Figure \ref{fig:cross} demonstrates the results of PCA into two dimensions, for a large proprietary and internally annotated dataset containing speech segments from multiple domains such as movies, TV series and interviews. Subfigures illustrate the distributions of the speech segments into the two PCA dimensions for three emotional classes: anger, happiness and sadness. In addition, we illustrate the decision surfaces for a simple kNN classifier. The results demonstrate how the  blue class (anger) is similarly distributed between the red and green (sadness and happiness respectively) for the two first domains (Series and Movies) in Fig~\ref{fig:cross:series} and Fig.~\ref{fig:cross:movies} respectively, based on the primary PCA dimension (x axis). On the other hand, for the interviews domain, the primary PCA dimension is not enough to discriminate between the emotional classes as we see in Fig.~\ref{fig:cross:interviews}. On the contrary, the anger and happiness classes are mostly discriminated based in the second PCA dimension. Interestingly, this unsupervised distribution is quite similar to the Valence-Arousal affective representation. This example demonstrates how an unsupervised dimensionality reduction can be very sensitive to changes in domain when illustrating emotional content.

\begin{figure}
\centering
\subfigure[Pattern Search MDS\label{fig:mds:2d}]{
    \includegraphics[width=.4\textwidth]{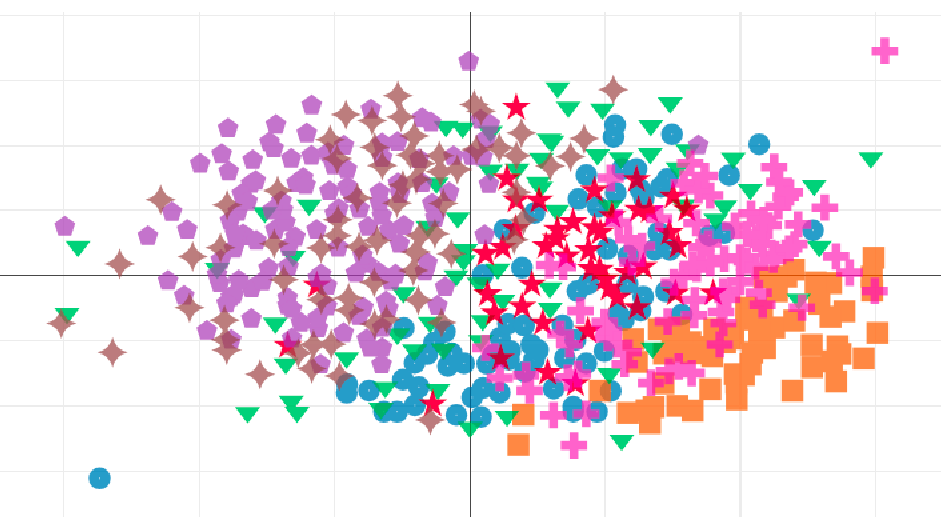}
}

\subfigure[LLE\label{fig:lle:2d}]{
    \includegraphics[width=.4\textwidth]{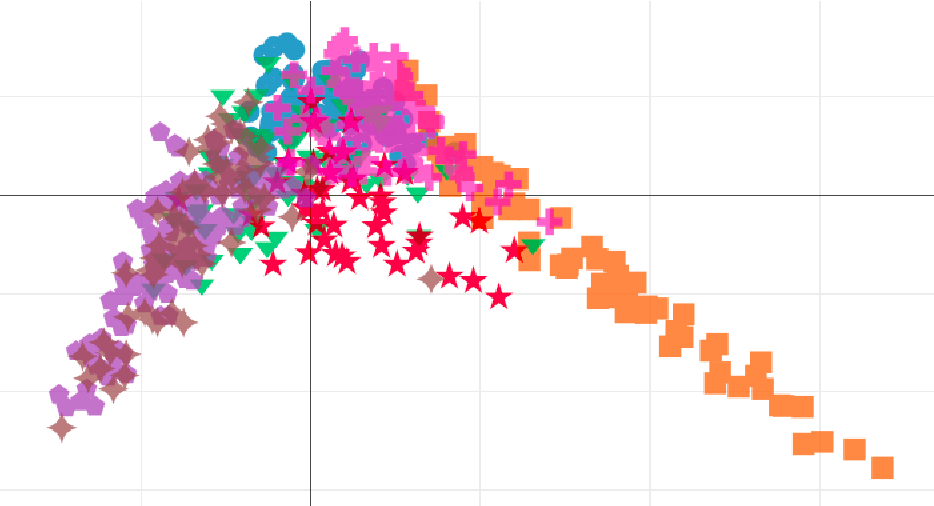}
}
\subfigure{\includegraphics[width=.9\linewidth]{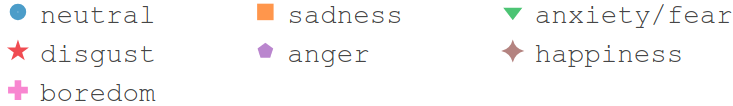}}

\caption{$2$D DR for fused feature set on Emo-DB}


\end{figure}

\begin{figure}
  \includegraphics[width=\linewidth]{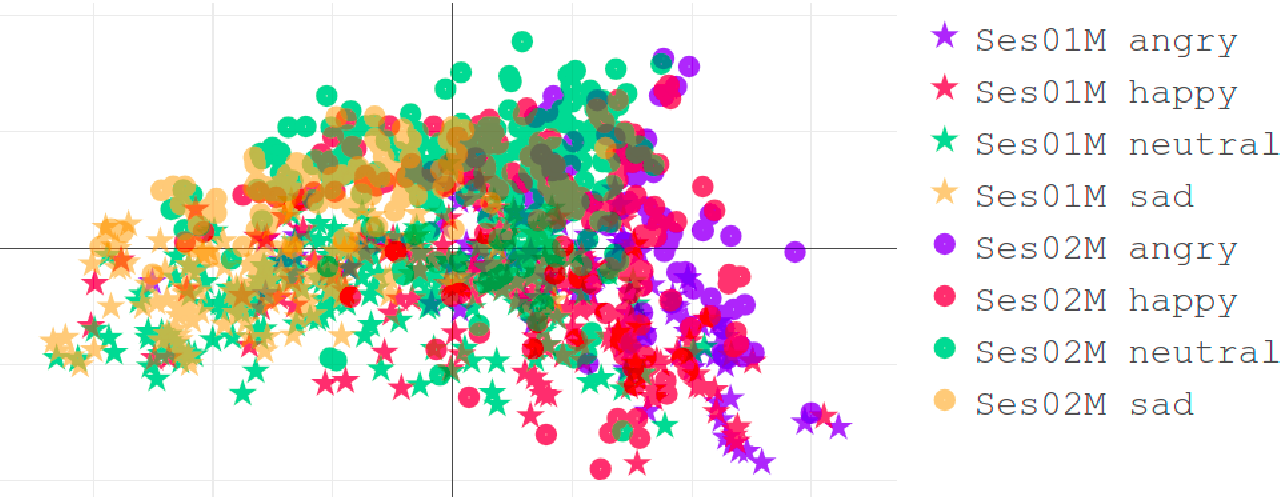}
  \caption{Isomap on fused features for 2 IEMOCAP speakers}
  \label{fig:isomap:speakers}
\end{figure}

Fig.~\ref{fig:mds:2d} shows the $2$D space created using Pattern Search MDS, which maps the points inside an elongated disk area. We can see on the left the anger points while the sadness points are on the right. Close to anger is happiness samples, while boredom is close to sadness. Other emotions lie in the middle. So it looks like that even in the $2$D space MDS learns meaningful representations, with $x$ axis being a latent feature that can encode arousal. On the contrary LLE, which tries to preserve local neighborhoods and yields poor recognition accuracy on the fused feature set concentrates most samples in the center as we can see in Fig.~\ref{fig:lle:2d}, but still we can observe low arousal emotions (sadness) being separated from high arousal ones (anger). In Fig.~\ref{fig:isomap:speakers} we show ISOMAP embeddings for two speakers in IEMOCAP. Observe, although ISOMAP cannot separate emotions, it achieves a better discrimination result, in terms of speaker separation, for this experiment.
One could consider basing a speaker diarizer on geodesic distances between samples. 

\begin{table}
\small
\centering
\caption{Classification on IS10 features for IEMOCAP (UA)}
\label{tab:res}
\begin{tabular}{c|cccc}
 & SVM (linear) & SVM (rbf) &  $k$NN & LR\\

 \hline
Pattern S. MDS & $\mathbf{56.0}$  & $57.5$ & $56.5$ & $55.4$ \\
SMACOF MDS & $55.8$ & $\mathbf{58.5}$  & $\mathbf{56.7}$ & $\mathbf{55.8}$ \\
PCA & $55.8$ & $57.7$ & $56.2$ & $\mathbf{55.8}$\\
ISOMAP & $52.3$  & $52.5$  & $51.7$ & $52.2$\\
LLE & $53.4$ & $54.2$ & $53.6$ & $53.2$\\
Modified LLE & $54.6$  & $47.0$ & $53.9$ & $55.5$\\
Spectral Emb. & $54.1$ & $54.3$ & $54.2$ & $55.1$\\
Autoencoder & $55.4$ & $57.8$ & $56.3$  & $55.5$\\
 \hline
 Original $1582$D &$54.7$ & $\mathbf{59.8}$ & $55.7$ & $\mathbf{56.9}$\\

\end{tabular}
\end{table}

\section{Conclusions}
\label{sec:conc}

In this work we explore the effects of unsupervised linear and non-linear DR on state-of-the-art speech features for SER. We evaluate these algorithms for speaker independent SER on IEMOCAP and Emo-DB. Experiments show that performance of low-rank representations is competitive to original high-dimensional representations. This phenomenon is hypothesized to be caused by the curse of dimensionality, since the number of samples in SER datasets does not span the high-dimensional space. Interpretation of results and vizualization of $2$D representations gives interesting insights on the high-dimensional structures. First insight is that IS10 features can be decomposed by use of linear DR, e.g. by use of PCA or MDS algorithms. Second, distance preserving DR can encode meaningful dimensions, e.g.  arousal. Third, speaker samples can be separated by isometric mappings. Fourth, unsupervised DR can be rather sensitive when illustrating cross-domain emotional content.  Future work will focus on creating end-to-end representations using autoencoders with distance preserving regularization and investigating the interesting insight on using geodesic-distance preserving representations for speaker separation.

\section{Acknowledgements}
\label{sec:ack}

This work has been partially supported by computational time granted from the Greek Research \& Technology Network (GRNET) in the National HPC facility - ARIS and the EU-IST H2020 BabyRobot project under grant \#687831.

\newpage
\bibliographystyle{IEEEtran}

\bibliography{refs}


\end{document}